\documentclass{article}
\usepackage[numbers,sort&compress]{natbib}
\usepackage{spconf,amsmath,graphicx}
\usepackage{array}
\usepackage{bm}
\usepackage{booktabs}
\usepackage{threeparttable}

\title{KeNet:Knowledge-enhanced Doc-Label Attention Network for Multi-label text classification}

\name{Bo Li $^{\spadesuit}$\textsuperscript{,\S{},*} \thanks{\S{} These authors contributed equally to this work.}\thanks{* The corresponding author.} \qquad Yuyan Chen $^{\clubsuit}$\textsuperscript{,\S{}} \qquad Liang Zeng $^{\spadesuit}$}
\address{$^{\spadesuit}$ Baidu Inc., Beijing, China \qquad
$^{\clubsuit}$ Fudan University, Shanghai, China}
\begin{document}
\ninept
\maketitle
\begin{abstract}
Multi-Label Text Classification (MLTC) is a fundamental task in the field of Natural Language Processing (NLP) that involves the assignment of multiple labels to a given text. MLTC has gained significant importance and has been widely applied in various domains such as topic recognition, recommendation systems, sentiment analysis, and information retrieval. However, traditional machine learning and Deep neural network have not yet addressed certain issues, such as the fact that some documents are brief but have a large number of labels and how to establish relationships between the labels. It is imperative to additionally acknowledge that the significance of knowledge is substantiated in the realm of MLTC. To address this issue, we provide a novel approach known as \textbf{K}nowledge-\textbf{e}nhanced Doc-Label Attention \textbf{Net}work (\textbf{\textit{KeNet}}). Specifically, we design an Attention Network that incorporates external knowledge, label embedding, and a comprehensive attention mechanism. In contrast to conventional methods, we use comprehensive representation of documents, knowledge and labels to predict all labels for each single text. Our approach has been validated by comprehensive research conducted on three multi-label datasets. Experimental results demonstrate that our method outperforms state-of-the-art MLTC method. Additionally, a case study is undertaken to illustrate the practical implementation of KeNet.

\end{abstract}
\vspace{-0.5em}
\section{Introduction}
\begin{figure}[t]
  \centering
  \includegraphics[width=0.9\linewidth]{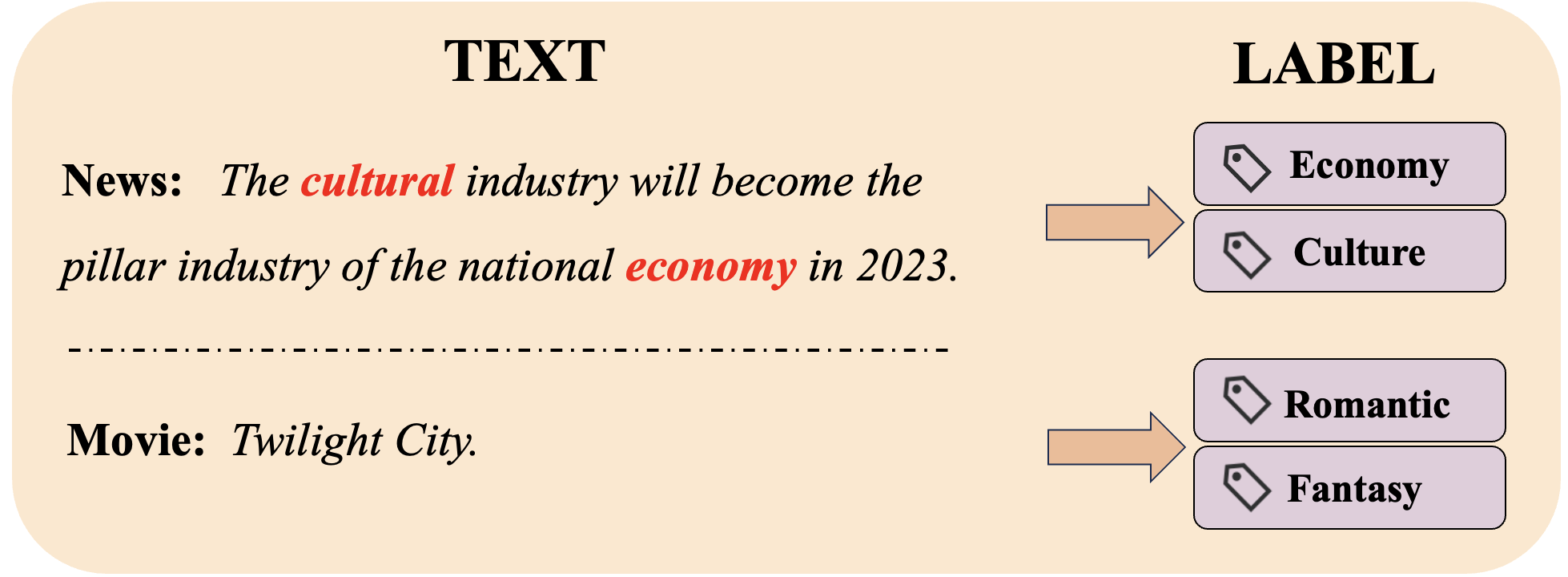}
  \caption{The example of Multi-label Text Classification.}
  \vspace{-2em} 
  \label{fig:example}
\end{figure}

\vspace{-0.5em}
\label{sec:intro}
MLTC aims to assign text input to multiple predetermined categories, and has been shown to be employed in several fields such as topic recognition, question answering, sentiment analysis, intention detection, tag recommendation and information retrieval\cite{yang2016hierarchical,kumar2016ask,cambria2014senticnet}. It permits the coexistence of numerous labels within a singular document, wherein each label signifies a distinct facet of the document's content. Therefore, it may be deduced that the entirety of semantic information encompassed inside a text consists of numerous interrelated or layered components. Figure \ref{fig:example} exemplifies the instances,the news story titled ``The cultural industry will become the pillar industry of the national economy in 2020'' may be categorized under either ``Economy'' or ``Culture''. Likewise, The movie ``Twilight City'' can be classified as ``romance'' and ``fantastic'' movie.

However, enormous difficulties impede our progress in solving the MLTC task accurately. Several difficult problems of MLTC can be summarized as follows: i) The number of labels for a given text is unknown, because some samples may have only one label while others may belong to dozens or even hundreds of topics; ii) The content of some documents is not rich enough to accurately predict the labels, because these documents belong to three or more labels.

MLTC methods can be broadly classified into two primary categories: traditional multi-label classification algorithms and deep learning-based algorithms. The traditional multi-label classification algorithms, such as BR\cite{Boutell04}, CC\cite{Read11}, LP\cite{Tsoumakas06}, ML-DT\cite{Clare01}, Rank SVM\cite{Elisseeff02} and ML-KNN\cite{Clare01}, are limited to capturing low-order correlations. There are challenges in the area of text representation of vector features, including limited expression capabilities and high cost of manual feature representation implementation. 
Deep learning-based algorithms, such as CNN and RNN\cite{Baker17,pal2020multi,ozmen2022multi,lu2023cnn}, are widely used in text classification, and with the introduction of transformer\cite{Bahdanau15,YangP18,Attentionxml,Attention_is_all,An_R-Transformer_BiLSTM_Model,liu2022co} and the pre-trained model such as BERT \cite{BERT-base}, significant improvements in classification performance have been achieved. However, they have also failed to capture high-order dependencies between labels or distinguish similar sub-labels, and the attention weights employed in these models lack the necessary strength to accurately assign all relevant labels to a given document.
Significantly, prior studies in the field of text categorization have not incorporated external knowledge, which has been demonstrated to hold significance\cite{sinoara2019knowledge,hong2022lea}.

\begin{figure*}[t]
  \centering
  \includegraphics[width=1\linewidth]{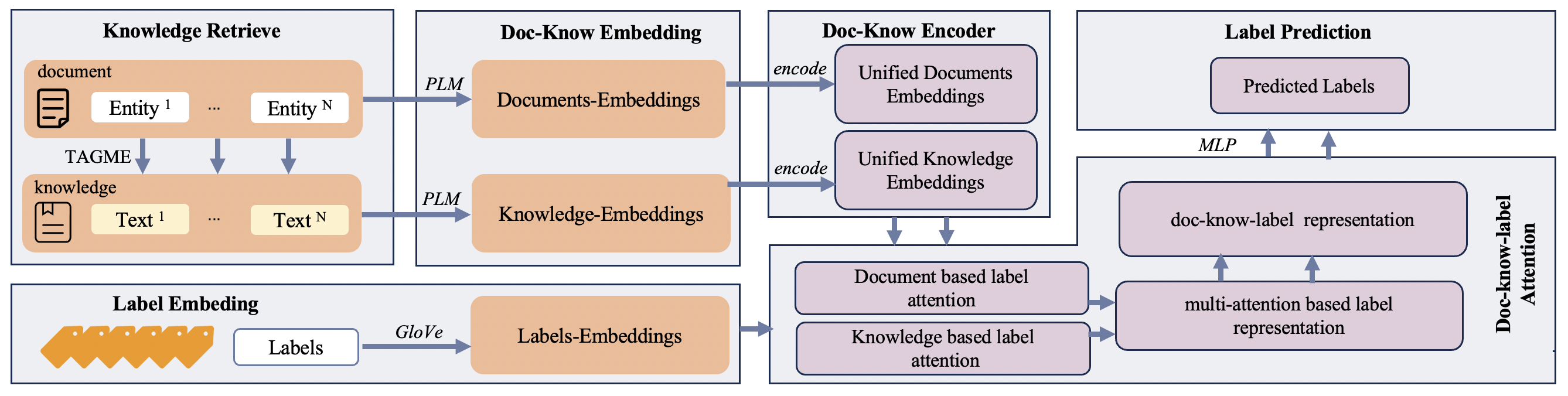}
  \caption{The architecture of the proposed KeNet model.}
  \vspace{-1.5em} 
  \label{fig:model}
\end{figure*}

To address the challenges mentioned above, we propose a novel Knowledge-enhanced Doc-Label Attention Network (KeNet) for MLTC, which aims at solving above-mentioned two issues including not rich documents and label correlation. The initial step is acquiring pertinent information to enhance the content of the documents. Subsequently, we make embeddings for documents and knowledge, and encode each of them as unified lengths. We also make embeddings for labels to capture the contextual relationship among each label set. In addition, we adopt attention mechanism for document-label pairs and knowledge-label pairs, respectively, and assign weights to obtain dependent label representation. Ultimately, we use a comprehensive representation of documents, knowledge and labels to predict all labels for each single text. 

The main contributions of this work can be summarized as follows. Firstly, we retrieve external knowledge based on entity linking techniques to provide richer information for documents. Secondly, we adopt attention mechanism for document-label pairs and knowledge-label pairs, respectively, to obtain a dependent label representation that contains comprehensive information between document, knowledge, and labels.  Ultimately, a series of experiments are performed on multiple real-world datasets. The results indicate that our proposed model outperforms all state-of-the-art MLTC models.

\vspace{-0.5em}
\section{Method}
\vspace{-0.5em}
In the paper, the structure of the model is shown in Figure ~\ref{fig:model}. It consists of six parts. Knowledge Retrieval refers to the process of accessing external knowledge in order to enhance the informational content of documents. Doc-know Embedding is to extract semantic features of documents and the corresponding knowledge. Doc-know Encoder is to obtain documents representation and knowledge representation of uniform length, respectively. Label Embedding is to map each label to a high-dimensional space to capture the interactive information of sub-labels. Doc-know-label Attention is to capture comprehensive features of document representation, knowledge representation and label embedding. Label Prediction is to make final multi-label text classification.

The MLTC task in this research can be summarized as a tuple set $S=\{(d_i,l_i)\}_{i=1}^N$ with $d_i$ and $l_i$ represents the $i$-th document denoted as $D=\{d_i | d_i=\{d^1,d^2,\cdots,d^n\}$ and its corresponding label sets denoted as $L=\{l_i | l_i=\{l^1,l^2,\cdots,l^m\}$. $N$, $n$ and $m$ are the total number of documents, the length of the $i$-th document and the number of labels of the $i$-th document, respectively. Our proposed KeNet model aims at assigning all suitable labels to its corresponding documents based on the conditional probability $Pr (l_i | d_i)$ to solve the MLTC task.
\vspace{-0.5em}
\subsection{Knowledge Retrieval}
\vspace{-0.5em}
\label{Entity_Descriptor}
Knowledge Retrieval is designed to introduce external knowledge to documents to provide richer information. It takes documents as input and outputs the unstructured text retrieved from the Wikipedia of the entities in the documents as the corresponding knowledge.
We first search the entities in the input document based on Wikipedia with the TAGME tool~\cite{TAGME}. 
We keep the entities whose confidence scores given by TAGME are larger than 0.5 in order to guarantee the accuracy of entities searching.
After that, for each retained entity, we crawl its Wikipedia text with the highest confidence.
If more than one entity in a document, we retrieve their specific Wikipedia texts and combine in the entity-order of the document. These combined texts are regarded as the knowledge of a document.
\vspace{-0.5em}
\subsection{Doc-know Embedding}
\vspace{-0.5em}
Doc-know Embedding is designed to capture the deep information of documents and knowledge. We use a Pre-trained Language Model (PLM, such as BERT~\cite{BERT-base}) to make embeddings for documents and knowledge, respectively. The input of PLM is documents and knowledge,
and the output is respective contextual representation of documents and knowledge. The process is shown as follows:
\begin{gather}
    Em^D = PLM(D),\quad Em^K = PLM(K)
\end{gather}
where $D$ and $K$ represent documents and knowledge, respectively.
\begin{table*}[t]
\caption{Comparisons of KeNet and fourteen baselines on RCV1-V2, AAPD, Reuters-21578. $(+)$ means the higher the value is, the better performance of the model. $(-)$ indicates the opposite.}
\label{tab:Comparsions1}
\begin{center}
\begin{threeparttable}
\scalebox{0.75}{
\setlength{\tabcolsep}{3.9mm}{
\begin{tabular}{l|cccc|cccc|cccc}
\toprule
\bf Datasets & \multicolumn{4}{c|}{\bf RCV1-V2}& \multicolumn{4}{c|}{\bf AAPD}& \multicolumn{4}{c}{\bf Reuters-21578} \\
\midrule
\bf Metrics & \bf HL(-)& \bf mP(+)& \bf mR(+)& \bf mF1(+)& \bf HL(-)& \bf mP(+)& \bf mR(+)& \bf mF1(+)& \bf HL(-)& \bf mP(+)& \bf mR(+)& \bf mF1(+)\\
\midrule
BR~\cite{Boutell04}&  0.0086& 0.904 & 0.816 & 0.858&  0.0316 &0.664 & 0.648 & 0.646&  0.0032 & 0.940 & 0.823 & 0.878\\
CC~\cite{Read11}& 0.0087& 0.887 & 0.828 & 0.857&  0.0306 &0.657 & 0.651 & 0.654&   0.0031 & 0.937 &  0.828 &  0.879\\
LP~\cite{Tsoumakas06}&  0.0087& 0.896 & 0.824 & 0.858&  0.0323 &0.662 & 0.608 & 0.634&-&-&-&-\\
\midrule
CNN~\cite{Yoon14}&0.0089&0.922 &  0.798 & 0.855&0.0256& 0.849 & 0.545 & 0.664&-&-&-&-\\
CNN-RNN~\cite{Guibin17}&0.0085 & 0.889 &0.825 & 0.856&0.0280& 0.718 & 0.618 & 0.664&0.0038 & 0.902 & 0.813 & 0.855\\
ML-Reasoner~\cite{A_novel_reasoning}& 0.0076& 0.889 &  0.860 &  0.870& 0.0248 &0.726 & 0.718 &  0.722&-&-&-&-\\
\midrule
S2S~\cite{Sutskever14}&0.0082&  0.883 & 0.849 & 0.866&0.0255 &0.743 & 0.646 & 0.691&-&-&-&-\\
S2S+Attn~\cite{Bahdanau15}&0.0081&  0.889 & 0.848 & 0.868&0.0261 &0.720 & 0.639 & 0.677&-&-&-&-\\
SGM~\cite{YangP18}&0.0075&  0.897 & 0.860 & 0.878&0.0245 &0.748 & 0.675 & 0.710&-&-&-&-\\
MDC~\cite{Lin18}& 0.0072& 0.891 &  0.873 &  0.882& 0.0240 &0.752 & 0.681 &  0.715&-&-&-&-\\
AttentionXML~\cite{Attentionxml}& 0.0079& 0.890 &  0.850 &  0.871& 0.0242 &0.757 & 0.685 &  0.715&-&-&-&-\\
Transformer~\cite{Attention_is_all}& 0.0072&  0.891 &  0.873 &  0.882& 0.0244 &0.744 & 0.676 &  0.698&-&-&-&-\\
LANRTN~\cite{An_R-Transformer_BiLSTM_Model}& 0.0070&  0.910 &  0.890 &  0.893& 0.0240 &0.762 & 0.689 &  0.718&-&-&-&-\\
HBLA~\cite{A_Hybrid_BERT}& 0.0063&  0.906 &  0.892 &0.899& 0.0223 &0.768 & 0.722 &  0.744&-&-&-&-\\
\midrule
\bf KeNet(ours)&\bf 0.0062&\bf 0.943&\bf  0.910&\bf 0.926&\bf 0.0221&\bf  0.845&\bf 0.698&\bf 0.764&\bf 0.0017 &\bf 0.988 & \bf 0.901 &\bf 0.942\\
\bottomrule
\end{tabular}
}}
\end{threeparttable}
\end{center}
\vspace{-2em} 
\end{table*}
\vspace{-0.5em}
\subsection{Doc-know Encoder}
\vspace{-0.5em}
Doc-know Encoder is designed to obtain unified contextual representation of given documents and knowledge. We adopt bidirectional LSTM to encode each of them and output $2H$-dimensional vectors. The unified representation of document and knowledge are denoted as $En^D \in R^{l_1 \times 2H}$ and $En^K \in R^{l_2 \times 2H}$ where $l_1$ and $l_2$ means the length of input documents and knowledge. The specific equations are shown as follows:
\begin{gather}
    En^D=\{[LSTM(\overrightarrow{Em^D_{t-1}},h_t); LSTM(\overleftarrow{Em^D_{t-1}},h_t)]\}_{t=1}^T\\
    En^K=\{[LSTM(\overrightarrow{Em^K_{t-1}},h_t); LSTM(\overleftarrow{Em^K_{t-1}},h_t)]\}_{t=1}^T
\end{gather}
\vspace{-0.5em}
\subsection{Label Embedding}
\vspace{-0.5em}
Label Embedding is designed to fully establish contextual relationship among labels for the reason that all labels for one document contain different but relevant semantic information. We convert the label set $L=\{l_i | l_i=\{l^1,l^2,\cdots,l^m\}$ for a document to embedding vectors $Em^L \in R^{M \times d}$ via GloVe \cite{Jeffrey14} with $M$ representing the total number of labels. The process is shown as follows:
\begin{gather}
    Em^L=GloVe(L)
\end{gather}
\vspace{-1.5em}
\subsection{Doc-know-label Attention}
\vspace{-0.5em}
Doc-know-label Attention is designed to capture interactive features between documents and their corresponding labels, as well as retrieved knowledge and their corresponding labels.
In the MLTC task, a single document belongs to several labels and a label can be attributed to several documents. Moreover, the corresponding knowledge is considered as supplementary information to enrich documents. 
Therefore, we adopt Doc-know-label attention which is the weighted attention of the document-label attention and the knowledge-label attention to fuse information between documents, knowledge and labels. Firstly, we apply self-attention mechanism on documents and the corresponding knowledge to obtain document attention $A_D$ and knowledge attention $A_K$. Based on the above two attention, we obtain independent document weight $\lambda_D$ and independent knowledge weight $\lambda_K$ which denotes contribution of documents in document-label pairs and knowledge in knowledge-label pairs, respectively:
\begin{gather}
    A_D=softmax(W_1^{'}tanh(W_1({En^D})^T))\\
    A_K=softmax(W_2^{'}tanh(W_2({En^k})^T))\\
    \lambda_D=\sigma((A_DEn^D)W_1^{''}),\quad \lambda_K=\sigma((A_KEn^K)W_2^{''})
\end{gather}
\begin{table}
  \caption{Statistics of three datasets.$W$, $N_{train}$, $N_{test}$ and $M$ denote the number of total words, training documents, test documents and total unique labels, respectively.}
  \label{tab:Statistics}
  \begin{center}
  \begin{threeparttable}
\scalebox{0.75}{
\setlength{\tabcolsep}{5.8mm}{
  \begin{tabular}{l|cccc}
    \toprule
    Dataset&$W$&$N_{train}$&$N_{test}$&$M$\\
    \midrule
    RCV1-V2 & 47,236 &23,149 &781,265 & 103\\
    AAPD & 69,399 &54,840 & 1,000 & 54\\
    Reuters-21578 & 18,637 &8,630 &2,158 & 90\\
  \bottomrule
\end{tabular}
}}
\end{threeparttable}
\end{center}
\vspace{-2.2em} 
\end{table}
Next, we apply document-label attention for document-label pairs to obtain document-based label attention $A_{L_D}$, and apply knowledge-label attention for knowledge-label pairs to obtain knowledge-based label attention $A_{L_K}$. Then we assign weight factors $\beta_1$ and $\beta_2$ ($\beta_1$+$\beta_2$=1) for each of two pairs to get multi-attention-based label representation $L^A$ and its independent label weight $\lambda_L$:
\begin{gather}
    A_{L_D}=(W_3Em^L)(W_3^{'}({En^D})^T)\\
    A_{L_K}=(W_4Em^L)(W_4^{'}({En^K})^T) \\
    L^A=\beta_1A_{L_D}En^D+\beta_2A_{L_K}En^K,\quad
    \lambda_L=\sigma(L^AW_5)
\end{gather}
After that, the final doc-know-label representation $S^A$ is calculated by multiplying dependent label weight $\lambda$:
\begin{gather}
    \lambda=\frac{\lambda_L}{\lambda_L+\lambda_D}+\frac{\lambda_L}{\lambda_L+\lambda_K},\quad
    S^A=\lambda L^A
\end{gather}
where $W_1$, $W_1^{'}$, $W_1^{''}$, $W_2$, $W_2^{'}$, $W_2^{''}$, $W_3$, $W_3^{'}$, $W_4$, $W_4^{'}$, $W_5$ are trainable parameters and $\sigma$ is sigmoid activation function (the same below).
\begin{table}[ht]
\caption{Ablation study of five derived models on RCV1-V2. KR: Knowledge Retrieval; DEm: Doc-know Embedding; DEn: Doc-know Encoder; LEm: Label Embedding; DA: Doc-know-label Attention.}
\label{tab:Ablation}
\begin{center}
\begin{threeparttable}
\scalebox{0.75}{
\setlength{\tabcolsep}{5.8mm}{
\begin{tabular}{l|cccc}
\toprule
\bf Metrics & \bf HL(-)& \bf mP(+)& \bf mR(+)& \bf mF1(+)\\
\midrule
w/o KR& 0.0085& 0.891& 0.868& 0.879\\
w/o DEm &0.0080&0.903 & 0.882 & 0.892 \\
w/o DEn&0.0067& 0.926 & 0.899 & 0.912 \\
w/o LEm&0.0079& 0.918 & 0.890 & 0.904 \\
w/o DA & 0.0083&0.895 &0.872 &0.883 \\
\midrule
\bf KeNet (ours)&\bf 0.0062&\bf 0.943&\bf 0.910&\bf 0.926\\
\bottomrule
\end{tabular}
}}
\end{threeparttable}
\end{center}
\vspace{-2.2em}
\end{table}
\vspace{-0.5em}
\subsection{Label Prediction}
\vspace{-0.5em}
Label Prediction is to use doc-know-label representation $S^A$ to make mMulti-Label Text Classification as follows:
\begin{gather}
\hat{y}=\sigma(W_p^{'}tanh(W_pS^A))
\end{gather}
where $W_p$, $W_p^{'}$ are trainable parameters.
We adopt cross-entropy loss as the loss function in our work which has been proved suitable for the MLTC task.
\vspace{-1.7em}
\section{Experiments}
\vspace{-0.5em}
We conduct experiments on three popular dataset, RCV1-V2, AAPD and Reuters-21578 and compare fourteen baselines to validate the performance of our proposed KeNet. We also make ablation study to analyze the effect of each module of KeNet. Furthermore, we make case study to further visualize the application of KeNet.
\vspace{-0.5em}
\subsection{Experimental setup}
\vspace{-0.5em}
We carry out our experiments on NVIDIA TESLA V100 GPU with Pytorch. We set the maximum length of each document and knowledge as 250, respectively, and adjust the embedding size of labels as 300. The dimension of hidden state in BiLSTM is set to 300. We use Adam optimizer with $\beta_1$ = 0.9 and $\beta_2$ = 0.999 in the training process. The batch size is adjusted to 128 and the learning rate is initialized to 1e-04. We evaluate model performance on test sets after 200 epochs with early stopping when the validation loss stops decreasing after 10 epochs. 

\textbf{Datasets, Baselines and Evaluation metrics.} In this research, we utilize three popular multi-label text datasets and the detailed statistics are shown in Table~\ref{tab:Statistics}. Based on investigation, the selected datasets are most used in previous research.
We compare our proposed KeNet with fourteen baselines as shown in Table~\ref{tab:Comparsions1}. The results of baselines are from their published paper.
Inspired by the previous work \cite{ZhangML07,Guibin17}, we apply Hamming Loss, micro-Precision, micro-Recall and micro-F1 which are also mostly used to validate the performance of KeNet.
\begin{figure}[t]
  \centering
  \includegraphics[width=0.9\linewidth]{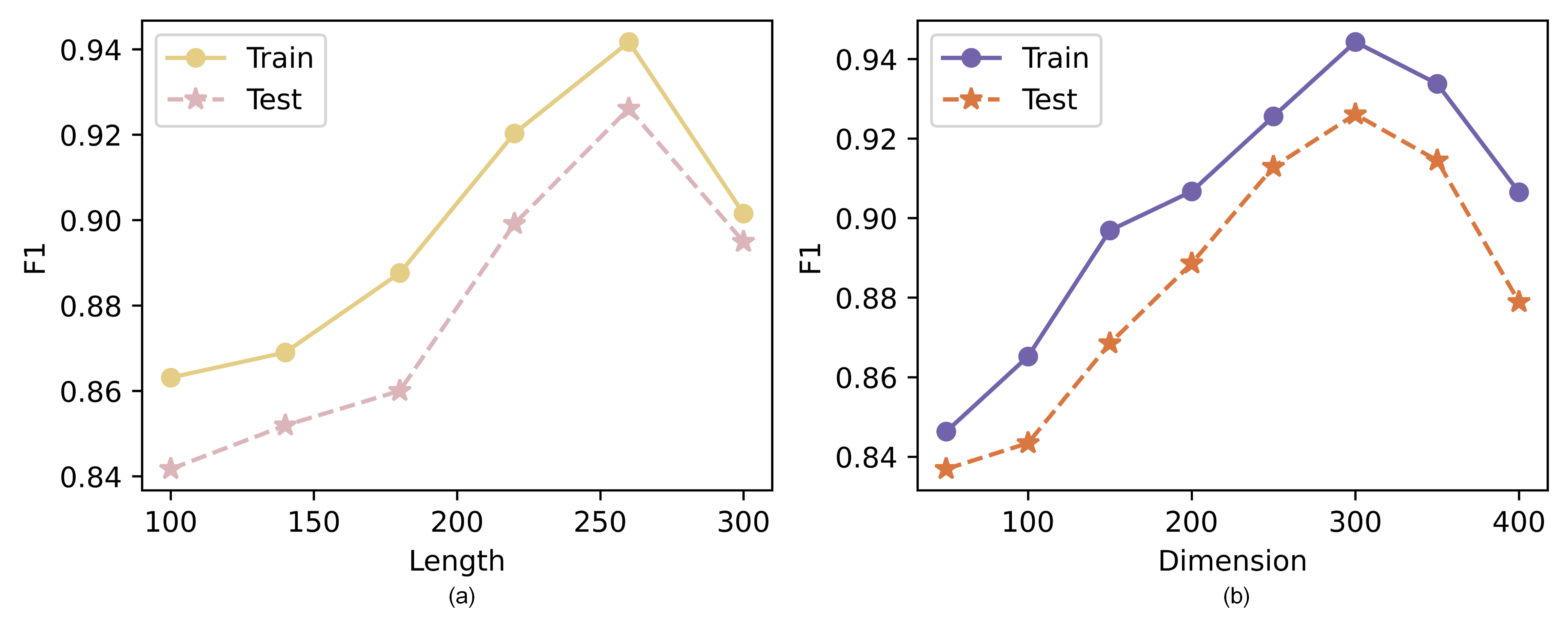}
  \caption{Influence of dimension of hidden state, and document and knowledge length on the RV1-V2 dataset.}
  \label{fig:Parameters}
  \vspace{-1.8em}
\end{figure}
\vspace{-0.5em}
\subsection{Main results}
\vspace{-0.5em}

Fourteen baselines can be divided into three categories, including machine learning methods (BR, CC, LP), conventional deep learning models (CNN, CNN-RNN) and Seq2Seq or attention-based approaches (S2S, S2S+Attn, SGM, MDC, AttentionXML, Transformer, LANRTN, HBLA). The results are shown in Table~\ref{tab:Comparsions1}.
We can find that generally conventional deep learning methods outperform machine learning models. It strongly demonstrates conventional deep learning model are superior in extracting deep semantic information than feature-engineering-driven traditional machine learning methods.
Moreover, The average results of Seq2Seq and attention-based approaches show an advantage over that of conventional deep learning models.
It indicates that Seq2Seq models and attention-based models are more capable of exploring latent label orders with global embedding.
Most importantly, the experiment results show that our proposed KeNet has the best performance on all three datasets, outperforming the current SOTA models on all metrics
(p$<$0.05 on student t-test for all above comparisons, and the same below).
The possible reason is that i) external knowledge provides richer information for documents; ii) PLM extract deeper and hierarchical contextual information of documents in Doc-know Embedding; iii) Label Embedding integrate all unique labels in order to capture latent connections between each label-pair; iv) Doc-know-label attention mechanism learn comprehensive information among  documents, knowledge and labels based on the dependent attention weights.
\vspace{-0.7em}
\subsection{Ablation study}
\vspace{-0.5em}
To analyze the contributions of each module of the proposed KeNet, we carry out ablation study of five derived models. We analyze results on RV1-V2 as shown in Table~\ref{tab:Ablation}.
The most important module is Knowledge Retrieval, which drops dramatically after removing it (w/o KR). It indicates that external knowledge is necessary supplement information for documents which can improve the performance of multi-label text classification. The second important one is Document-knowledge-label Attention, which also drops at a large degree after removing it (w/o DA). It demonstrates its function in extracting comprehensive semantic features among documents, knowledge and their corresponding labels. When we remove Doc-know Embedding (w/o DEm) and Label Embeddings (LEm), respectively, the drop degree is similar, which proves the embedding technique can effectively capture global semantic information of texts. Finally, the performance drops the least when we remove Doc-know Encoder (w/o DEn). It indicates that document and knowledge encoder further enhance semantic information of texts.
Above all, each component of the proposed KeNet has indispensable effect and the organic combination of these modules make contributions to KeNet's SOTA performance compared with baselines.
\begin{figure}[t]
  \centering
  \includegraphics[width=0.8\linewidth]{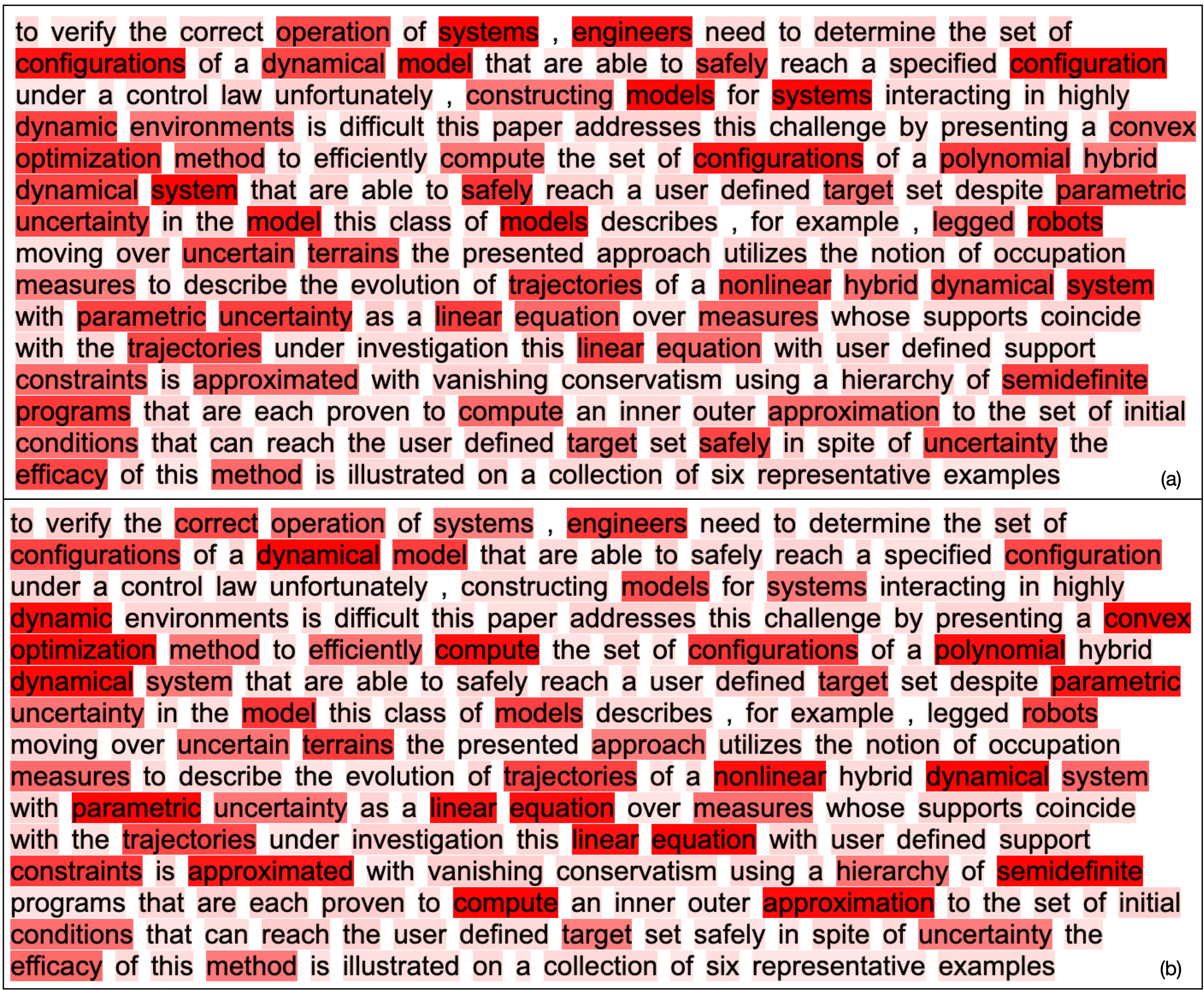}
  \caption{Visual analysis of KeNet on a MLTC task with label $cs.sy$ (a) and $math.oc$ (b).}
  \label{fig:casestudy1}
\end{figure}
\begin{figure}
  \centering
  \includegraphics[width=0.8\linewidth]{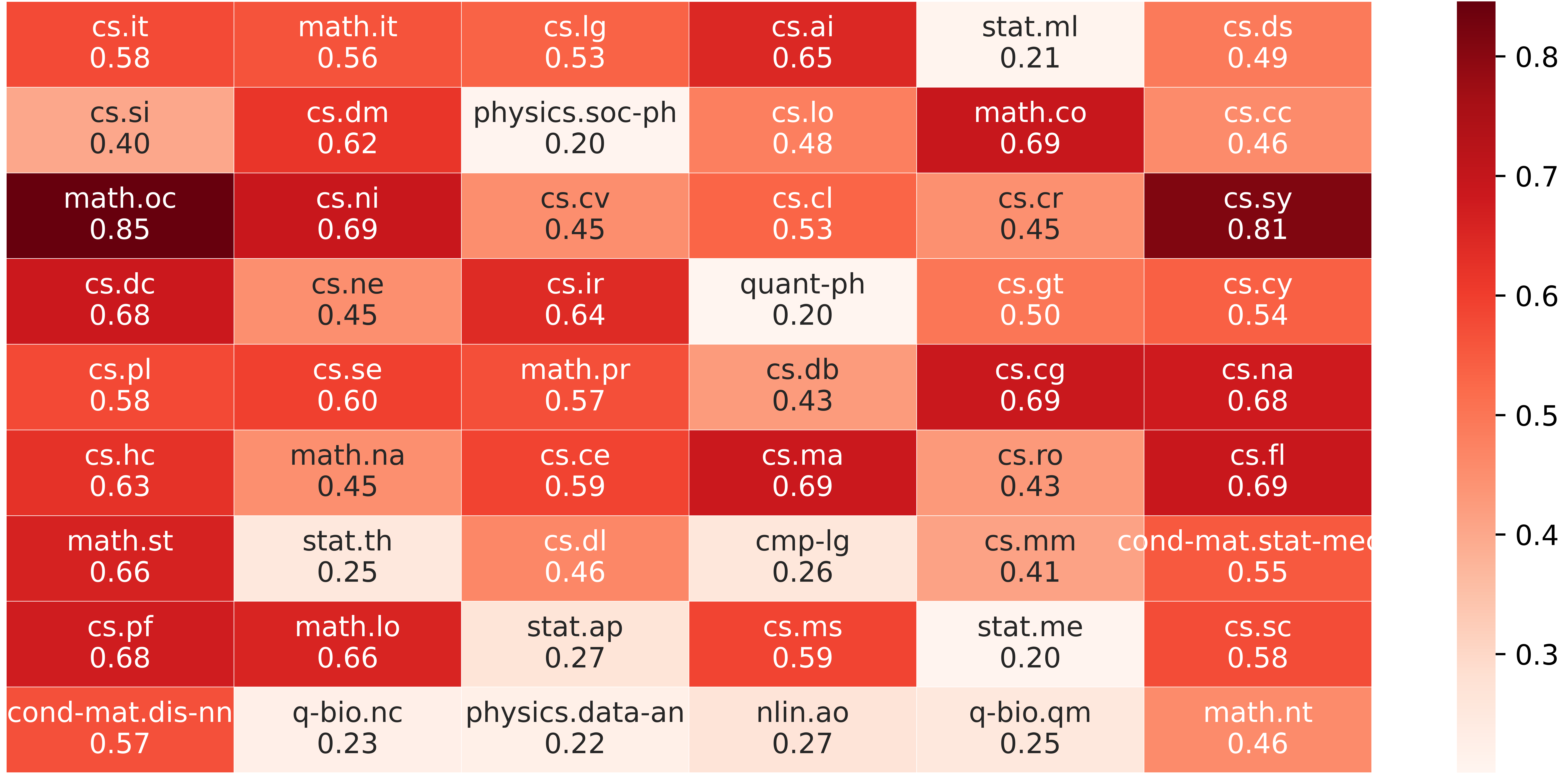}
  \caption{Weights of all labels of a given document.}
  \label{fig:casestudy2}
   \vspace{-1.8em}
\end{figure}
\vspace{-0.5em}
\subsection{Parameters sensitivity}
\vspace{-0.5em}
We also carry out experiments on the length of input documents in Doc-know Embedding and the dimension of hidden state in Doc-know Encoder. Results on the RV1-V2 dataset are shown in Figure~\ref{fig:Parameters}.
We can find that the peaks of length (see Figure~\ref{fig:Parameters} (a)) are 250 both on the training set and test set. 
Next, the peaks of dimension are 300 (see Figure~\ref{fig:Parameters} (b)) both on the training set and test set. 
After or before peaks, the performance all drops, which
indicates that Doc-know Embedding and Doc-know Encoder manage to capture more significant information within acceptable limits.

\vspace{-0.5em}
\subsection{A Case study}
\vspace{-0.5em}
We conduct a case study to better understand how to classify multi-label documents using KeNet. We visualize a document which is labeled $cs.sy$ and $math.oc$ shown in Figure~\ref{fig:casestudy1}. First, we aim to explore contributions of each word in the whole document based on labels $cs.sy$ and $math.oc$, respectively. For example, words of high contribution in the document for the first label $cs.sy$ are $systems$, $engineers$, etc., which are covered with deep red. They have a positive effect on predicting the target label $cs.sy$, which is in line with human intuition. Similar to the second label $math.oc$.Next, we reveal probabilities of all unique labels calculated by KeNet through a heatmap shown in Figure~\ref{fig:casestudy2}. The probabilities of correct labels $cs.sy$ and $math.oc$ obtain 0.85 and 0.88 which substantially exceed other labels. 
\vspace{-1.5em}
\section{Conclusion}
\vspace{-0.5em}
In this paper, we propose a novel Knowledge-enhanced Doc-Label Attention Network named KeNet, which designed to reliably predict all labels associated with each text. KeNet incorporates external knowledge, a thorough attention mechanism, and four additional modules. Experimental results on three datasets and four common evaluation metrics demonstrate that our proposed model outperforms all state-of-the-art MLTC models. We also carry out case study to visualize its real applications. In the future, we will generalize our model with more datasets to increase its robustness and extend its applications to more scenarios.
\clearpage
\label{sec:refs}
\bibliographystyle{IEEEbib}
\bibliography{strings,refs}
\end{document}